\documentclass[letterpaper]{article} % DO NOT CHANGE THIS
\usepackage{aaai25}  % DO NOT CHANGE THIS
\usepackage{times}  % DO NOT CHANGE THIS
\usepackage{helvet}  % DO NOT CHANGE THIS
\usepackage{courier}  % DO NOT CHANGE THIS
\usepackage[hyphens]{url}  % DO NOT CHANGE THIS
\usepackage{graphicx} % DO NOT CHANGE THIS
\usepackage{natbib}  % DO NOT CHANGE THIS AND DO NOT ADD ANY OPTIONS TO IT
\usepackage{caption} % DO NOT CHANGE THIS AND DO NOT ADD ANY OPTIONS TO IT
\usepackage{algorithm}
\usepackage{algorithmic}

\usepackage{newfloat}
\usepackage{listings}
\DeclareCaptionStyle{ruled}{labelfont=normalfont,labelsep=colon,strut=off} % DO NOT CHANGE THIS
\floatstyle{ruled}
\newfloat{listing}{tb}{lst}{}
\floatname{listing}{Listing}
\usepackage{listings}
\usepackage{xcolor}
\usepackage{tcolorbox}
\usepackage{pifont}
\usepackage{amssymb}
\usepackage{amsfonts}
\usepackage{subcaption}
\usepackage{makecell}
\usepackage{multirow}
\usepackage{graphicx}
\usepackage{booktabs}
\usepackage{amsmath}
\usepackage{cleveref}
\usepackage{xcolor}
\DeclareMathOperator*{\argmax}{arg\,max}

\title{What Are Step-Level Reward Models Rewarding?\\Counterintuitive Findings from MCTS-boosted Mathematical Reasoning}

\author{
  Yiran Ma\textsuperscript{\rm 1}\thanks{These authors contributed equally. Work was done during their internships at TAL Education Group.},
  Zui Chen\textsuperscript{\rm 2}\footnotemark[1],
  Tianqiao Liu\textsuperscript{\rm 3},
  Mi Tian\textsuperscript{\rm 3},
  Zhuo Liu\textsuperscript{\rm 4},
  Zitao Liu\textsuperscript{\rm 5}\thanks{Zitao Liu is the corresponding author.},
  Weiqi Luo\textsuperscript{\rm 5}
}
\affiliations{
  \textsuperscript{\rm 1}Zhejiang University, Hangzhou, China\\
  \textsuperscript{\rm 2}ShanghaiTech University, Shanghai, China\\
  \textsuperscript{\rm 3}TAL Education Group, Beijing, China\\
  \textsuperscript{\rm 4}University of Rochester, New York, USA\\
  \textsuperscript{\rm 5}Jinan University, Guangzhou, China\\
  mayiran@zju.edu.cn, chenzui@shanghaitech.edu.cn, \{liutianqiao1, tianmi\}@tal.com, zhuo.liu@rochester.edu, \\
  \{liuzitao, lwq\}@jnu.edu.cn
}

\usepackage{bibentry}
\begin{document}

\maketitle

\begin{abstract}
  Step-level reward models (SRMs) can significantly enhance mathematical reasoning performance through process supervision or step-level preference alignment based on reinforcement learning. The performance of SRMs is pivotal, as they serve as critical guidelines, ensuring that each step in the reasoning process is aligned with desired outcomes. Recently, AlphaZero-like methods, where Monte Carlo Tree Search (MCTS) is employed for automatic step-level preference annotation, have proven particularly effective.  However, the precise mechanisms behind the success of SRMs remain largely unexplored. To address this gap, this study delves into the counterintuitive aspects of SRMs, particularly focusing on MCTS-based approaches. Our findings reveal that the removal of natural language descriptions of thought processes has minimal impact on the efficacy of SRMs. Furthermore, we demonstrate that SRMs are adept at assessing the complex logical coherence present in mathematical language while having difficulty in natural language. These insights provide a nuanced understanding of the core elements that drive effective step-level reward modeling in mathematical reasoning. By shedding light on these mechanisms, this study offers valuable guidance for developing more efficient and streamlined SRMs, which can be achieved by focusing on the crucial parts of mathematical reasoning.
\end{abstract}

% Uncomment the following to link to your code, datasets, an extended version, or similar.

\section{Introduction}
\begin{figure}[ht!]
  \includegraphics[width=\linewidth]{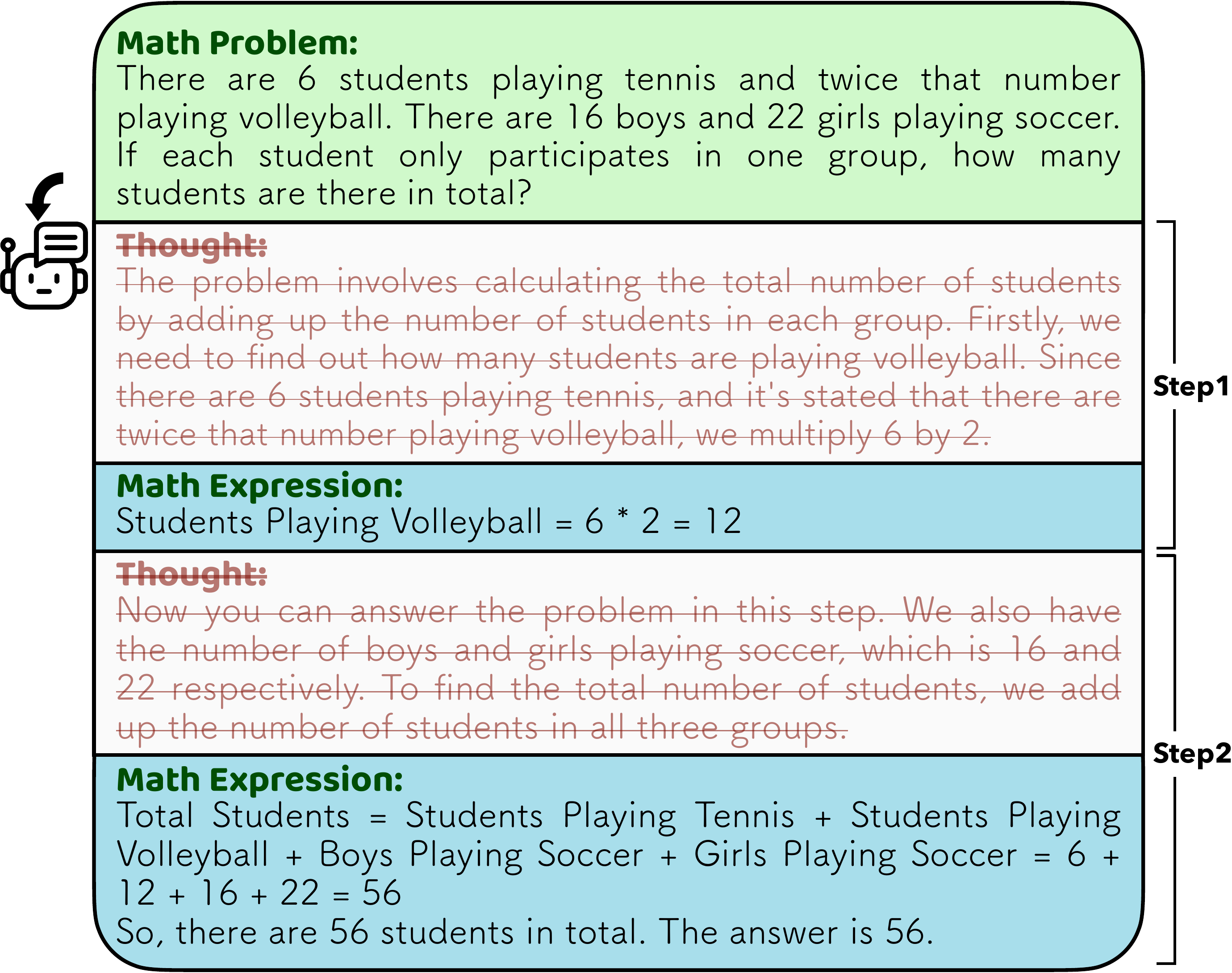}
  \caption{Each step in an LLM’s process of solving mathematical problems can be divided into the thought process and the execution of corresponding calculations. We find that natural language descriptions of the thought processes are not essential for step-level reward modeling.}
  \label{fig:teaser}
\end{figure}
Large Language Models (LLMs) have demonstrated their remarkable capabilities across a wide range of tasks, such as information extraction, natural language understanding, etc \cite{zhao2023survey}, totally revolutionizing the deep learning community. Among these capabilities, reasoning stands out as a critical area of focus, especially mathematical reasoning, which needs to be further improved due to its complex nature. Numerous studies have shown that multi-step reasoning often facilitated through Chain-of-Thought (CoT) prompting, can significantly enhance model performance on reasoning tasks \cite{zhou2022least,besta2024graph,ding2023everything,yao2024tree,wang2022self,wei2022chain,zheng2024automatic,li2024enhancing,zhan2024knowledge}.

Recently, guided tree-search methods further improved reasoning performance by exploring various reasoning paths through online simulation to identify the optimal solution paths \cite{hao2023reasoning, hao2024llm,feng2023alphazero}. Although a better reasoning path leads to a better performance, the length of these reasoning chains leads to an exponential increase in the search space, resulting in substantial computational costs. Given the high expense of LLM inference, performing an online tree search for each reasoning problem introduces repeated and unnecessary overhead.

To address this issue, step-level reward models (SRM) was proposed to improve search efficiency. \citet{lightman2023let} introduced the process reward model (PRM), which employs human-annotated step-level scores for reward modeling, and \citet{ma2023let} further demonstrated the effectiveness of SRMs in math reasoning and coding tasks. Then, Math-Shepherd \cite{wang2023math}, systematically generates step-level preference data through exhaustive reasoning process traversal to train reward models and reinforce the model's capabilities. More recently, inspired by AlphaZero, Monte Carlo Tree Search (MCTS) \cite{xie2024monte, chen2024alphamath, chen2024step} was then used for collecting preferences more efficiently because of its capability of balancing exploration and exploitation. These trained SRMs can effectively enhance reasoning performance by either assisting step-level preference alignment with proximal policy optimization (PPO) during training stage or serving as step verifiers during inference stage.

Despite the significant achievements in mathematical reasoning performance achieved by the SRMs constructed by MCTS-based method, the exact workings of these reward models and what they are truly rewarding remain unclear. Brain and cognitive scientists have argued that diverse thinking and reasoning processes do not necessarily rely on natural language. \cite{fedorenko2024language}.
A skilled human mathematician, for instance, can determine whether a mathematical expression is logically coherent and numerically correct without the participation of the natural language.
Building on this idea, our research explores a similar hypothesis for LLMs: that \textbf{natural language descriptions of thought processes are not essential for mathematical reasoning within these models.}
We suppose that LLMs can be trained to recognize preferences for mathematical language directly during problem-solving, without relying on natural language descriptions.
This implies that LLMs might be capable of understanding and processing mathematical reasoning through the intrinsic structure of mathematical language, potentially leading to more efficient and focused training methods that bypass the need for natural language explanations.
Furthermore, it is believed that incorrect solutions often arise from wrong mathematical calculations or logical errors \cite{zhang2024accessing}, with the latter being more challenging \cite{chen2024alphamath}.
Therefore, we further investigate the effectiveness of SRMs in evaluating logical coherence in pure mathematical language, demonstrating that the improvements are not merely the result of encouraging correct calculations within a single step.
Additionally, and somewhat surprisingly, we found that SRMs struggle to learn how to evaluate logical coherence in natural language.
This will further support that natural language is not necessary for step-level reward modeling.

% \TODO{Math language: single/link}
\begin{figure*}[t!]
  \centering
  \includegraphics[width=1.0\linewidth]{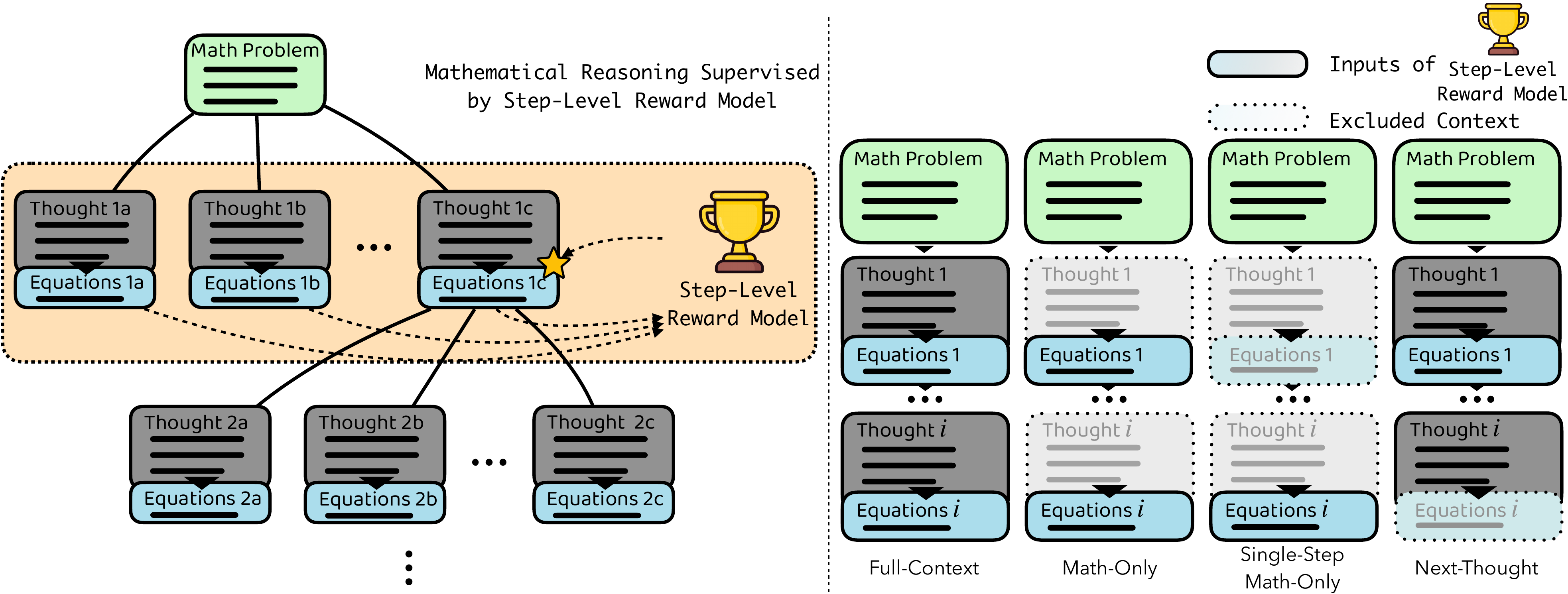}
  \caption{Illustration of the role of SRMs in mathematical reasoning and the SRMs with different input structures we investigate.}
  \label{fig:main}
\end{figure*}

To investigate the respective roles of natural language and mathematical language in step-level reward modeling, we decompose each step of the reasoning path into two components: natural language descriptions of thought processes and math expressions (\Cref{fig:teaser}). The ablation studies are conducted by selectively removing different parts from the inputs of the SRMs. This decomposition mirrors the human problem-solving process in mathematics, which typically involves an initial phase of thinking through the problem, followed by the execution of calculations based on that thought process. The thought processes include the strategy to be taken in that step, while the calculations are the executions of the thought processes.
In other words, our decomposition aims to separate the natural language (composing the `thoughts') from the mathematical expressions (contained in the execution of `thoughts').
This framework aims to foster a deeper understanding of the role of natural language for step-level reward modeling.

To summarize, our experiments support that SRMs appear to have some intrinsic affinity for mathematical expression, not natural language. Specifically, we propose the following key insights.
\begin{enumerate}
  \item Natural language descriptions of thought processes are not necessary for successful step-level reward modeling.
  \item SRMs not only promote accurate calculations within individual steps but also effectively assess the challenging logical coherence in mathematical language.
  \item Assessing logical coherence in natural language is difficult, and SRMs often struggle with this task.
\end{enumerate}

\section{Preliminaries}
\subsection{Markov Decision Process}
\subsubsection{Definition}
A Markov Decision Process (MDP) is a mathematical framework used to model decision-making problems. This framework is fundamental for addressing a wide range of reinforcement learning (RL) problems where the outcomes are partially random and partially controllable. An MDP is defined by a tuple \((S, A, P, R, \gamma)\), where:
\begin{itemize}
  \item \(S\) is the set of states.
  \item \(A\) is the set of actions.
  \item \(P\) is the transition probability function, \(P(s_{t+1}|s_t, a_t)\), which defines the probability of transitioning to state \(s_{t+1}\) given the current state \(s_t\) and action \(a_t\).
  \item \(R\) is the reward function, \(R(s_t, a_t, s_{t+1})\), which defines the reward received after transitioning from state \(s_t\) to state \(s_{t+1}\) by taking action \(a_t\).
  \item \(\gamma\) is the discount factor, which determines the importance of future rewards.
\end{itemize}

\subsubsection{Bellman Expectation Equation}
\label{sec:bellman}
For \textbf{state value function} \(V(s)\), the Bellman Expectation Equation is:
\[
  V^\pi(s) = \mathbb{E}_{a \sim \pi(\cdot | s)} \left[ \mathbb{E}_{s' \sim P(\cdot | s, a)} \left[ R(s, a, s') + V^\pi(s') \right] \right]
\]
For \textbf{state-action value function} \(Q(s, a)\), the Bellman Expectation is:
% \begin{equation*}
% \begin{aligned}
%     Q^\pi(s, a) = \mathbb{E}_{s' \sim P(\cdot | s, a)} \left[ R(s, a, s') +
%     \mathbb{E}_{a' \sim \pi(\cdot | s')} \left[ Q^\pi(s', a') \right] \right]
% \end{aligned}
% \end{equation*}
{\small
  \begin{equation*}
    \begin{aligned}
      Q^\pi(s, a) = \mathbb{E}_{s' \sim P(\cdot | s, a)} \left[ R(s, a, s') +
      \mathbb{E}_{a' \sim \pi(\cdot | s')} \left[ Q^\pi(s', a') \right] \right]
    \end{aligned}
  \end{equation*}
}
% \begin{equation*}
% \begin{aligned}
% Q^\pi(s, a) = \mathbb{E}_{s' \sim P(\cdot | s, a)} \Big[ R(s, a, s') + \\
% \mathbb{E}_{a' \sim \pi(\cdot | s')} \left[ Q^\pi(s', a') \right] \Big]
% \end{aligned}
% \end{equation*}

\subsubsection{Optimal Value Functions}
The optimal value functions are defined as:
\begin{equation}
  \label{eq:optimal_value}
  \begin{aligned}
    V^*(s)    & = \max_{\pi} V_\pi(s)    \\
    Q^*(s, a) & = \max_{\pi} Q_\pi(s, a)
  \end{aligned}
\end{equation}
Therefore, the relationship between the optimal value functions and the Bellman Optimality Equation is:
\begin{equation}
  V^*(s) = \max_a Q^*(s, a)
\end{equation}

\section{Setup}
\subsection{LLM's Math Reasoning as MDP: Our Definition}
\label{sec:mdp_reasoning}
\Cref{fig:main} shows the mathematical reasoning process with each step decomposed into thought and math expressions. Specifically, our MDP definition is as follows:
\[
  \text{MDP} = (S, A, P, R)
\]
where:
\begin{itemize}
  \item \textbf{State} The state space \(S\) consists of states defined as \(s_i = (T_k, E_k)_{k=0}^i\), representing a sequence of thoughts \(T_k\) and equations \(E_k\) up to step \(i\).
  \item \textbf{Action} The action space \(A\) consists of actions defined as \(a_i = T_{i+1}\), representing the natural language descriptions of the subsequent thought proposed by the LLM.
  \item \textbf{State Transition} \(P(s_{i+1} | s_i, a_i)\) is the state transition function, defining the probability of transitioning to state \(s_{i+1}\) from state \(s_i\) after taking action \(a_i\). This function is implemented by the LLM generating the corresponding math expression \(E_{i+1}\) based on the next thought \(a_i=T_{i+1}\) and the current state \(s_i = (T_k, E_k)_{k=0}^i\).
  \item \textbf{Reward Function} \(R(s_i, a_i, s_{i+1})\) is the reward function, defining the immediate reward received after transitioning to state \(s_{i+1}=(T_k, E_k)_{k=0}^{i+1}\) from state \(s_i\) by taking action \(a_i\). We define the reward up to state \(s_{i+1}\) based on whether it can lead to the correct final answer:
    \begin{equation}
      \label{eq:reward}
      R(s_i, a_i, s_{i+1}) =
      \begin{cases}
        1, & \text{final answer is correct} \\
        0, & \text{final answer is incorrect}
      \end{cases}
    \end{equation}
\end{itemize}
Additionally, policy \(\pi(a_i | s_i)\) is implemented by the LLM generating the thought of the next step \(a_i=T_{i+1}\) based on the current state \(s_i = (T_k, E_k)_{k=0}^i\). According to \Cref{eq:optimal_value}, the goal of an agent is to maximize \(V_{\pi}(s_i)\) or \(Q_{\pi}(s_i, a)\) by generating the correct thoughts \(T\) in each step.

In summary, a language model plays a dual role in the MDP framework:
\begin{enumerate}
  \item \textbf{As an Agent} The LLM is responsible for making decisions by selecting appropriate actions (next thoughts \(T_{i+1}\)) at each state, following the policy \(\pi(a_i | s_i)\).
  \item \textbf{As a World Model} The LLM also acts as the world model \(P(s_{i+1} | s_i, a_i)\) by predicting action outcomes (state transitions) using its internal knowledge and training data. It simulates the environment of mathematical reasoning by executing thought \(T_{i+1}\) through corresponding calculations, thus providing the prediction of new states \(s_{i+1}\).
\end{enumerate}

\subsection{MCTS for Step-Level Preference Collection}
Understanding the natural correspondence between math reasoning and MDP, we can readily use MCTS for efficient step-level preference collection. The MCTS starts from a root node \(s_0\), which is a math problem in mathematical reasoning tasks. Then, each new node corresponds to a state update. Each iteration of MCTS can be divided into four phases: Selection, Expansion, Rollout, and Back-propagation.

\begin{enumerate}
  \item \textbf{Selection.} The selection phase in MCTS involves traversing the tree from the root node \(s_0\) (the initial math problem) to a leaf node using a selection policy. This policy, typically the Upper Confidence Bound for Trees (UCT) formula, balances exploration and exploitation. At node \(s_i\), the next node is chosen by:
    \begin{equation}
      s_{i+1}^* = \arg\max_{s_{i+1}} \left[ \frac{c(s_{i+1})}{N(s_{i+1})} + w_{\text{exp}} \cdot \sqrt{\frac{\log N(s_i)}{N(s_{i+1})}} \right],
    \end{equation}
    where \(c(s_{i+1})\) is the correct counts, \(N(s_i)\) and \(N(s_{i+1})\) are visit counts, and \(w_{\text{exp}}\) balances exploration and exploitation. This process continues until an unexplored node is found.
  \item \textbf{Expansion.} Upon reaching a leaf node, $n$ new candidate actions (thoughts) \(\{a_i^j \mid j=1,...,n\}\) are generated by the agent given the current state \(s_i\). Given the candidate actions (thoughts), the world model will execute them through mathematical calculations, constructing the new candidate states \(\{s_i^j \mid j=1,...,n\}\). These candidate states are added as child nodes to the current node to expand the tree, allowing for a broader exploration of potential problem-solving paths.
  \item \textbf{Rollout.} The rollout phase simulates the reasoning process from the newly expanded node to a terminal state or predefined maximum depth. The score of a node is then obtained according to \Cref{eq:reward}. This procedure estimates the scores of the new nodes according to the simulation results, informing the back-propagation phase.
  \item \textbf{Back-propagation.} Results from the rollout are propagated back up the tree to update values and visit counts of each node. Starting from the final state, the effectiveness of the problem-solving process updates the value \(V(s)\) of each state. This procedure improves the selection policy for future iterations.
\end{enumerate}
After completing MCTS, step-level preference pairs can be gathered by comparing the values of the nodes in each tree.

\subsection{Step-level Reward Modeling}
\label{sec:srms}
After collecting all the preference pairs, step-level reward models can be constructed through contrastive learning. Based on our MDP definition, an SRM is regarded as the action-value function \(Q(s,a)\) or the value function \(V(s)\).
Specifically, we investigate different reward models for ablation studies, where reward models take different inputs to evaluate the ongoing reasoning process. Accordingly, we define four reward models (\Cref{fig:main}-right) for the ablation study:
% \TODO{reorder}
\begin{itemize}
  \item \textbf{Full-Context Step-level Reward Model (FC-SRM)} This model takes both the thoughts and math expressions of the current state as input.
    \begin{equation}
      V_{1}(s_{i})=V_1((T_{k}, E_{k})_{k=0}^{i})
    \end{equation}
  \item \textbf{Math-Only Step-level Reward Model (MO-SRM)} This model takes only the math expressions of the current state as input, excluding the natural language descriptions of thought processes.
    \begin{equation}
      V_{2}(s_{i})=V_2((E_{k})_{k=0}^{i})
    \end{equation}
  \item \textbf{Single-Step Math-Only Step-level Reward Model (SSMO-SRM)} This model takes only the newest math expression of the ongoing reasoning process as input, excluding the natural language and all the previous math expressions.
    \begin{equation}
      V_{3}(s_{i})=V_3(E_{i})
    \end{equation}
  \item \textbf{Next-Thought Step-level Reward Model (NT-SRM)} This model takes both the thoughts and math expressions of the current state as input, and evaluates the next thought. According to our definition, the next thought is the action taken by the agent. Thus this reward model is the action-value function under our MDP definition of mathemetical reasoning.
    \begin{equation}
      Q(s_i,a_i)=Q((T_{k}, E_{k})_{k=0}^i, T_{i+1})
    \end{equation}
\end{itemize}

\subsection{Beam Search with Step-Level Reward Model}\label{sec:bs}
Given the SRMs trained on the preference data, it is commonly used for step-level preference alignment to update the policy. The purpose of this procedure is to generate the best action through the updated policy \(\pi'\), thereby reducing the overhead caused by online MCTS. It is also possible to update the world model \(P\) with these preference pairs as better accuracy indicates better mathematical performance.

\begin{algorithm}
  \caption{Beam Search Algorithm}
  \label{alg:beam_search}
  \begin{algorithmic}[1]
    \REQUIRE Initial state $s_0$, beam size $B$, candidate count $c$
    \STATE Initialize beam $\mathcal{B} \gets \{s_0\}$
    \WHILE{$\mathcal{B}$ is not empty}
    \STATE Initialize empty list $\mathcal{B}_{\text{next}} \gets \emptyset$
    \FOR{each state $s_i$ in $\mathcal{B}$}
    \STATE Generate a set of candidate actions $\{a^1_i, a^2_i, \dots, a^c_i\}$ based on $s_i$
    \FOR{each action $a^j_i$ in $\{a^1_i, a^2_i, \dots, a^c_i\}$}
    \STATE Compute the next state $s^j_{i+1} \gets P(s_{i+1} | s_i, a^j_i)$
    \STATE Evaluate the score of $s^j_{i+1}$
    \STATE Add $s^j_{i+1}$ to $\mathcal{B}_{\text{next}}$
    \ENDFOR
    \ENDFOR
    \STATE Sort $\mathcal{B}_{\text{next}}$ by score and keep the top $B$ states
    \STATE Update beam $\mathcal{B} \gets \text{top } B \text{ states from } \mathcal{B}_{\text{next}}$
    \ENDWHILE
    \RETURN the best state from the final beam
  \end{algorithmic}
\end{algorithm}

As this study focuses on the SRMs, our experiments will not include the preference alignment procedure.
Instead, we can use the SRMs as the scoring function during beam search (BS) \Cref{alg:beam_search} for simplification.
This simplification excludes potential uncertainties in the alignment process, providing a more straightforward understanding of SRMs’ effectiveness.
\textbf{Notably, setting $B = 1$ makes BS effectively become greedy search (GS).}

The greedy search can be regarded as a reasoning process supervised by an SRM (\Cref{fig:main}-left).
Indeed, with an infinite number of samples, the optimal actions and states identified through the policy \(\pi\) and the world model \(P\) will converge to the optimal actions and states similar to those generated by the optimal policy \(\pi^*\) in \Cref{eq:optimal_value}, respectively.
\begin{equation}
  \lim_{n\rightarrow \infty} P(\argmax_{\{a_{t}\}_{t=0}^n} Q(s,a_t)=\arg\max_{a\in A_{\pi}(s)}Q(s,a)) = 1
\end{equation}
where \(a_t\sim\pi(a|s)\) and \(A_{\pi}(s)\) denotes the state space of actions generated by the policy \(\pi\) given state \(s\).
Similarly, for states, we also have
\begin{equation}
  \lim_{n\rightarrow \infty} P(\argmax_{\{s'_{t}\}_{t=0}^n} V(s'_t)=\argmax_{s'\in S(s,a)}V(s')) = 1
\end{equation}
where \(s_t \sim \mathbb{E}_{a_{t-1}\in \pi(a|s_{t-1})}P(s|s_{t-1},a_{t-1})\).

% The next best action is
% \begin{equation}
%     a = \arg\max_{a} Q_{\pi}(s, a),
% \end{equation}
% and the next best state is
% \begin{equation}
%     s = \arg\max_{s} V_{\pi}(s)
% \end{equation}
% Therefore, we can use the reward models to score the actions or the states generated by the policy \(\pi\) and world model \(P\).

% 原来的主实验表格，包含了2个base
\begin{table*}[ht!]
  \centering
  % \resizebox{\linewidth}{!}{%
  \begin{tabular}{lllllcc}
    \toprule
    \textbf{Agent \& World Model}               & \multirow{2}{*}{\parbox[t]{1.3cm}{\textbf{Historical \newline Thoughts}}} & \multirow{2}{*}{\parbox[t]{1.3cm}{\textbf{Historical \newline Equations}}} & \multirow{2}{*}{\parbox[t]{1.3cm}{\textbf{Next \newline Thoughts}}} & \multirow{2}{*}{\parbox[t]{1.3cm}{\textbf{Next \newline Equations}}}& \multicolumn{2}{c}{\textbf{Accuracy (Gain) \%}}                 \\
    \cmidrule{1-1}\cmidrule{6-7}
    Llama-3-8B-Instruct &                                                 &                                                  &                                           &                                            & \textbf{GSM8K}                               & \textbf{MATH}          \\
    \midrule
    Pass@1 (3-shots)                   &                                                 &                                                  &                                           &                                            & 78.47 (+0.00)                       & 31.16 (+0.00) \\
    \midrule
    %  & \parbox[t]{1.5cm}{Historical \newline Thoughts} & \parbox[t]{1.5cm}{Historical \newline Equations} & \parbox[t]{1.5cm}{Next \newline Thoughts} & \parbox[t]{1.5cm}{Next \newline Equations} &                                     &               \\
    % \cmidrule{2-5}
    +GS w/ SRM (DeepSeek-Math-7B-Base) &   &   &   &   &                                     &               \\
    \hspace{1cm}Full-Context SRM                    & \ding{51}                                       & \ding{51}                                        & \ding{51}                                 & \ding{51}                                  & 86.20 (+7.73)                       & 38.58 (+7.42) \\
    \hspace{1cm}Math-Only SRM                      & \ding{55}                                       & \ding{51}                                        & \ding{55}                                 & \ding{51}                                  & 85.82 (+7.35)                       & 39.64 (+8.48) \\
    \hspace{1cm}Single-Step Math-Only SRM                   & \ding{55}                                       & \ding{55}                                        & \ding{55}                                 & \ding{51}                                  & 82.11 (+3.64)                       & 37.46 (+6.30) \\

    \hspace{1cm}Next-Though SRM                      & \ding{51}                                       & \ding{51}                                        & \ding{51}                                 & \ding{55}                                  & 79.38 (+0.91)                       & 30.98 (-0.18) \\
    \midrule
    +GS w/ SRM (Qwen2-7B)              &                                                 &                                                  &                                           &                                            &                                     &               \\
    \hspace{1cm}Full-Context SRM                     & \ding{51}                                       & \ding{51}                                        & \ding{51}                                 & \ding{51}                                  & 82.94 (+4.47)                       & 35.58 (+4.42) \\
    \hspace{1cm}Math-Only SRM                     & \ding{55}                                       & \ding{51}                                        & \ding{55}                                 & \ding{51}                                  & 83.78 (+5.31)                       & 35.10 (+3.94) \\
    \hspace{1cm}Single-Step Math-Only SRM                   & \ding{55}                                       & \ding{55}                                        & \ding{55}                                 & \ding{51}                                  & 81.65 (+3.18)                       & 33.08 (+1.92) \\
    \hspace{1cm}Next-Though SRM                      & \ding{51}                                       & \ding{51}                                        & \ding{51}                                 & \ding{55}                                  & 81.73 (+3.26)                       & 31.40 (+0.24) \\
    \bottomrule
  \end{tabular}%
  % }
  \caption{SRMs act as step-level scoring functions during GS. Sample \(c=5\) candidates of the subsequent step at each node and use beam size \(B=1\) (greedy search). The agent and the environment model is Llama-3-8B-Instruct. The reward models are trained based on Deepseek-Math-7B-Base or Qwen2-7B.}
  \label{tab:main}
\end{table*}
\section{Experiments}
\subsection{Implementation Details}
\subsubsection{Datasets} To construct step-level preference pairs through MCTS, we use the math problems and their corresponding final answers from the training data of GSM8K \cite{cobbe2021training} and MATH \cite{hendrycks2021measuring}. The accuracies are evaluated on the test data.

\subsubsection{Models} The reasoning process is conducted by the dialogue between two LLMs. We use the Llama-3-8B-Instruct \cite{metaintrollama3} as both the agent and world model in MCTS because of its excellent ability to follow instructions.

\subsubsection{Prompt} One LLM (as agent) is instructed to generate natural language descriptions of thoughts, and the other (as world model) is instructed to execute the thoughts. For specific prompts, see Appendix.
\subsubsection{Baseline} We use Llama-3-8B-Instruct construct the `Pass@1' baseline based on our prompt with 3 shots.
\subsubsection{MCTS for Step-Level Preference Collection} The MCTS requires the agent sampling \(n=6\) candidate actions at each expansion phase and iterates \(500\) times on each problem to evaluate the quality of each node. Notably, to avoid the influence of the variation of answer format, we use a supervised fine-tuned (SFT) model based on DeepSeek-Math-7B-Base to assert the correctness of the solution after each rollout during the search. This model is also used in our evaluation pipeline. To strengthen the preferences, only the preference pairs whose difference of value is greater than 0.7 are assumed valid. For detailed hyperparameters, see Appendix.
\subsubsection{Reward Training} DeekSeek-Math-7B-Base \cite{shao2024deepseekmath} or Qwen2-7B \cite{yang2024qwen2} is used as the base model for SRM training. Each SRM is trained on two instances, with each instance equipped with 8 A800 GPUs. For detailed hyperparameters, see Appendix.
\begin{figure*}[ht!]
  \centering
  \begin{subfigure}[b]{0.95\linewidth}
    \centering
    \includegraphics[width=\linewidth]{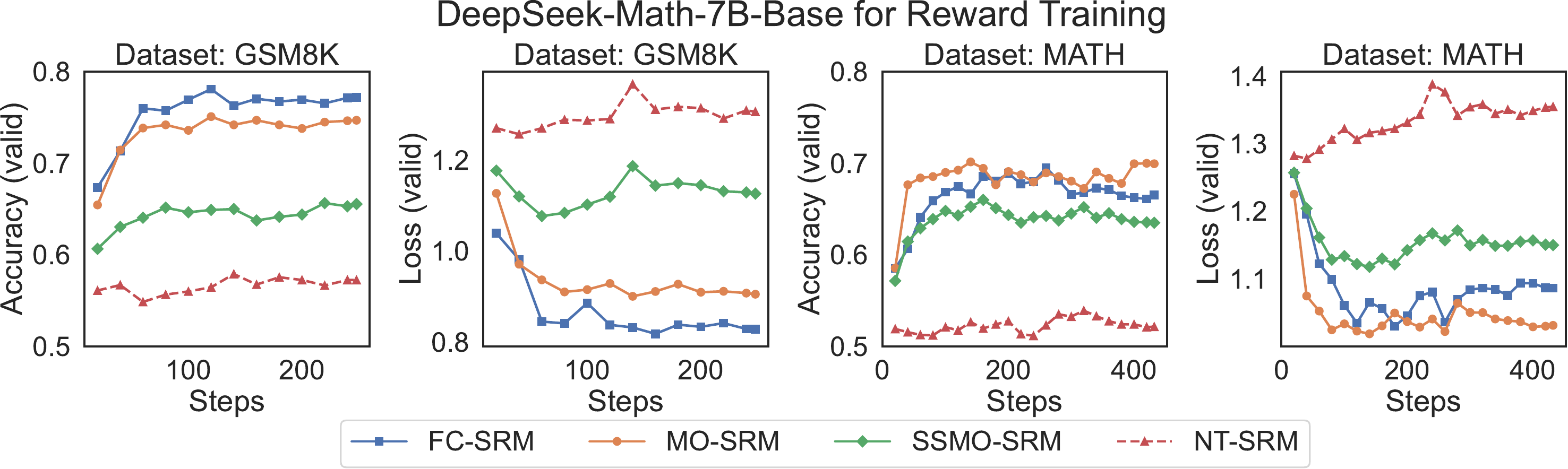}
    % \caption{Input Format: $V_2((E_{k})_{k=0}^{i})$}
    \label{fig:deepseek_curves}
  \end{subfigure}\\
  \begin{subfigure}[b]{0.95\linewidth}
    \centering
    \includegraphics[width=\linewidth]{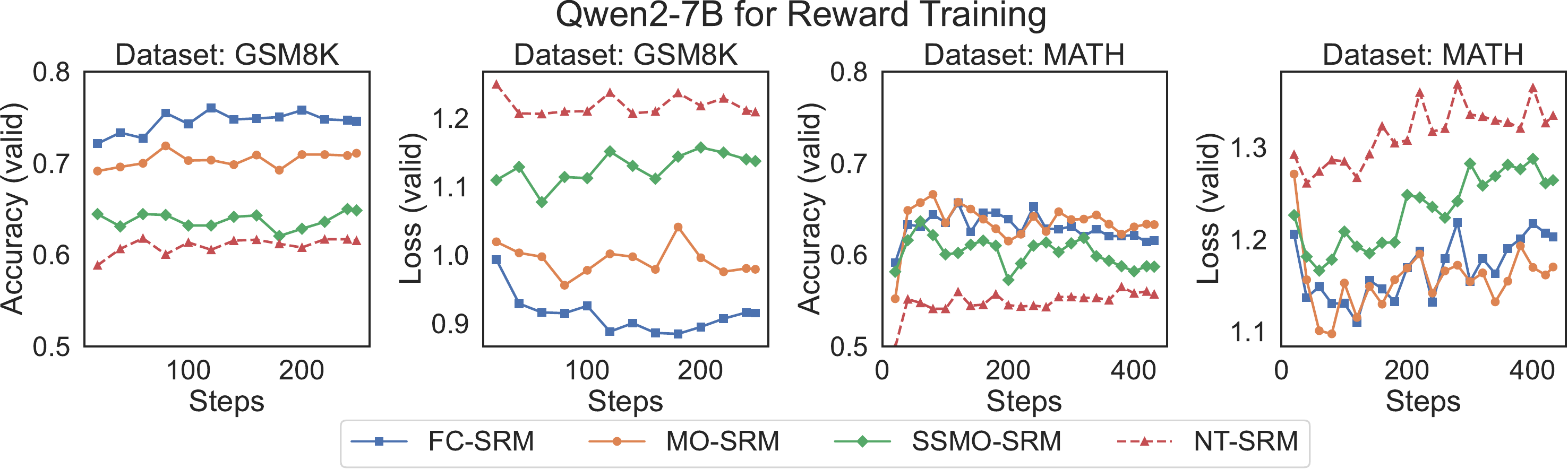}
    % \caption{Input Format: $V_1((T_{k}, E_{k})_{k=0}^{i})$}
    \label{fig:qwen_curves}
  \end{subfigure}
  \caption{Effect of natural language descriptions and math expressions on step-level reward modeling. The agent and the environment model is Llama-3-8B-Instruct. The reward models are trained based on Qwen2-7B or Deepseek-Math-7B-Base. (Note that the `accuracy' here is the accuracy of preference during reward training.)}
  \label{fig:training_curves}
\end{figure*}

\subsection{Main Results}
After collecting all the step-level preference pairs through MCTS, datasets are constructed for FC-SRM, MO-SRM, SSMO-SRM, and NT-SRM training by selecting the corresponding components in each piece of data. The training curves are shown in \Cref{fig:training_curves}. These SRMs are subsequently used as scoring functions in greedy search, the accuracy and absolute gains over baseline are reported in \Cref{tab:main}.
The analyses will be included in the following sections.

\subsection{Do we really need natural language?}
Intuitively, one might expect that natural language descriptions provide essential contextual information and aid SRMs' cognitive understanding. The SRMs with different input formats: full-context (FC) and math-only (MO) are trained to investigate this aspect.

\begin{figure}[H]
  \centering
  \includegraphics[width=\linewidth]{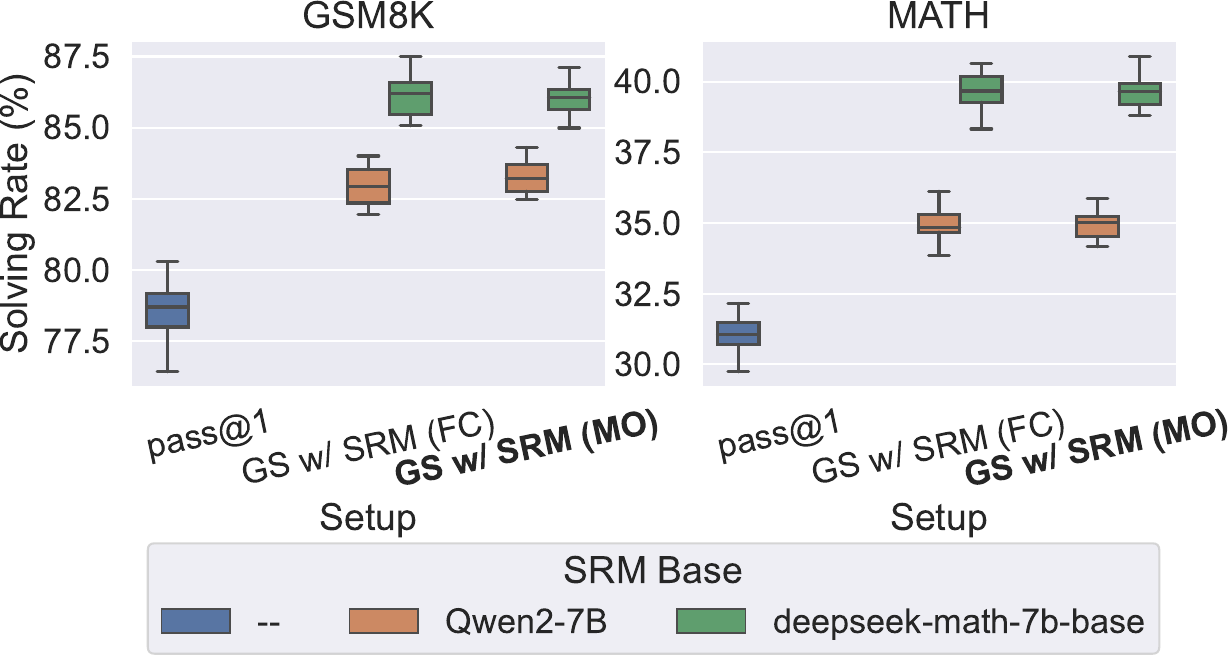}
  \caption{SRMs take only mathematical expressions as input demonstrate the same ability during the greedy search as those take full context as input. The boxplot is obtained through 20 runs over the dataset.}
  \label{fig:nl_boxplot}
\end{figure}

\subsubsection{Removing natural language has a minimal effect on step-level reward modeling.}
FC-SRMs and MO-SRMs exhibit very similar performance in both preference prediction accuracy and greedy search, suggesting that successful step-level reward modeling is not contingent upon natural language descriptions, which is contrary to intuition.
Even without the natural language descriptions of thoughts at each step, the MO-SRMs can still be successfully trained (\Cref{fig:training_curves}).
\Cref{tab:main} and \Cref{fig:nl_boxplot} further show the performance of these SRMs when used as scoring functions during greedy search. In setups such as MATH with DeekSeek-Math-7B-Base as the base model of SRM, the MO-SRM (39.64\%) can even outperform the FC-SRM (38.58\%).
We further conducted t-tests to provide a more detailed statistical comparison between the FC-SRMs and MO-SRMs across different datasets and base models. For the GSM8K dataset, the t-test results are \( t = -0.18 \), \( p = 0.86 \) for Qwen2-7B, and \( t = -0.14 \), \( p = 0.89 \) for deepseek-math-7b-base. For the MATH dataset, the results are \( t = 0.79 \), \( p = 0.44 \) for Qwen2-7B, and \( t = 0.77 \), \( p = 0.45 \) for deepseek-math-7b-base. In all cases, the p-values are greater than 0.05, indicating that the differences in performance between the FC-SRM and MO-SRM are not statistically significant. These results support the conclusion that omitting natural language from the inputs of SRMs has negligible effects on the effectiveness of SRMs.
% , which is consistent with the findings in previous studies.

% Moreover, the RMs \(Q((T_{k}, E_{k})_{k=0}^i, T_{i+1})\) do not work well. These observations all suggest that the LLMs as RMs have difficulties being trained to evaluate the preferences in the form of natural language. This should make us rethink the role of the natural language in MCTS-boosted math reasoning.
% \begin{figure*}[htp]
%     \centering
%     \begin{subfigure}{0.45\linewidth}
%         \centering
%         \includegraphics[width=\linewidth]{figures/gsm8k_ss.png}
%         % \caption{Input Format: $V_1((T_{k}, E_{k})_{k=0}^{i})$}
%         \label{fig:gsm8k_single}
%     \end{subfigure}
%     % \hfill
%     \begin{subfigure}{0.45\linewidth}
%         \centering
%         \includegraphics[width=\linewidth]{figures/math_ss.png}
%         % \caption{Input Format: $V_2((E_{k})_{k=0}^{i})$}
%         \label{fig:math_single}
%     \end{subfigure}
%     \caption{MO-SRM outperforms SSMO-SRM.}
%     \label{fig:coherence_effect}
% \end{figure*}

\subsection{Can SRMs evaluate logical coherence in math language?}
The success of MCTS-based methods is attributed to the ability to avoid logical and numerical errors. It is commonly believed that logical errors are more difficult to evaluate, while MCTS-based methods are believed a competitive solution to this challenge by collecting such preferences.
In this section, we investigate the role of natural language and mathematical language in assessing the logical coherence included in pure mathematical language by comparing SSMO-SRM, MO-SRM, and NT-SRM.

Specifically, if the contextual information in the input of an SRM is useful, its performance should surpass that of SSMO-SRM, which takes only the current step as input.
This ability is referred to as the model's capacity to assess logical coherence, meaning it can determine whether a subsequent step logically follows from the information and conclusions derived in the previous context. The results are shown in \Cref{tab:main}.

\subsubsection{LLMs can be trained to evaluate logical coherence in pure mathematical language.}
For DeepSeek-Math-7B-Base, MO-SRM achieves an accuracy gain of +7.35\% on GSM8K and +8.48\% on MATH, which is higher than the gains +3.64\% and 6.30\% observed for SSMO-SRM.
Similarly, for Qwen2-7B, MO-SRM achieves an accuracy gain of +5.31\% on GSM8K and +3.94\% on MATH, higher than that of SSMO-SRM +3.18\% and +1.92\%.
This substantial difference indicates that MO-SRM, which considers the full sequence of mathematical expressions, is effective at capturing logical coherence, rather than only focusing on the current step. This finding indicates that logical coherence in mathematical language can be assessed by LLMs as SRMs.

\subsubsection{The SRMs have difficulties being trained to evaluate the logical coherence in the form of natural language.}
Based on our MDP definition, even after the mathematical expressions are stripped away from the current reasoning step, the natural language descriptions still include the details of the actions to be executed. In other words, the SRMs should be able to learn from these constructed preferences to identify which actions are useful for problem-solving.
However, as shown in \Cref{fig:training_curves}, the dashed curves illustrate the challenges in training NT-SRMs, which were designed to evaluate the quality of the next thoughts. The training processes across various datasets and base models consistently demonstrate the difficulty in identifying preferences based solely on the descriptions of thoughts during reward training. The results presented in \Cref{tab:main} further highlight the poor performance of NT-SRMs when used as scoring functions. These findings suggest that the implicit logic conveyed through natural language is difficult for LLMs to capture and evaluate effectively.

\subsection{Additional Analysis}
\begin{table}[H]
  \centering
  % \resizebox{\linewidth}{!}{%
  \begin{tabular}{lcc}
    \toprule
    \textbf{Agent \& World Model}&  \multicolumn{2}{c}{\textbf{Accuracy (Gain) \%}}\\
    \cmidrule{2-3}
    Llama-3-70B-Instruct & \textbf{GSM8K}&\textbf{MATH}\\
    \midrule
    Pass@1 (3-shots) & \makecell{90.37\\(+0.00)} & \makecell{48.48\\(+0.00)}\\
    % \midrule
    +GS /w MO-SRM\footnotemark[1]&  \makecell{92.95\\(+2.58)}& \makecell{54.12\\(+5.64)}\\
    \bottomrule
  \end{tabular}
  % }
  \caption{Supervise a larger model (Llama-3-70B-Instruct).}
  \label{tab:llama3_70b}
\end{table}
\footnotetext[1]{The MO-SRM here is trained based on DeepSeek-Math-7B-Base with preference data generated through MCTS performed by Llama-3-8B-Instruct.}

\begin{figure}[H]
  \centering
  \includegraphics[width=\linewidth]{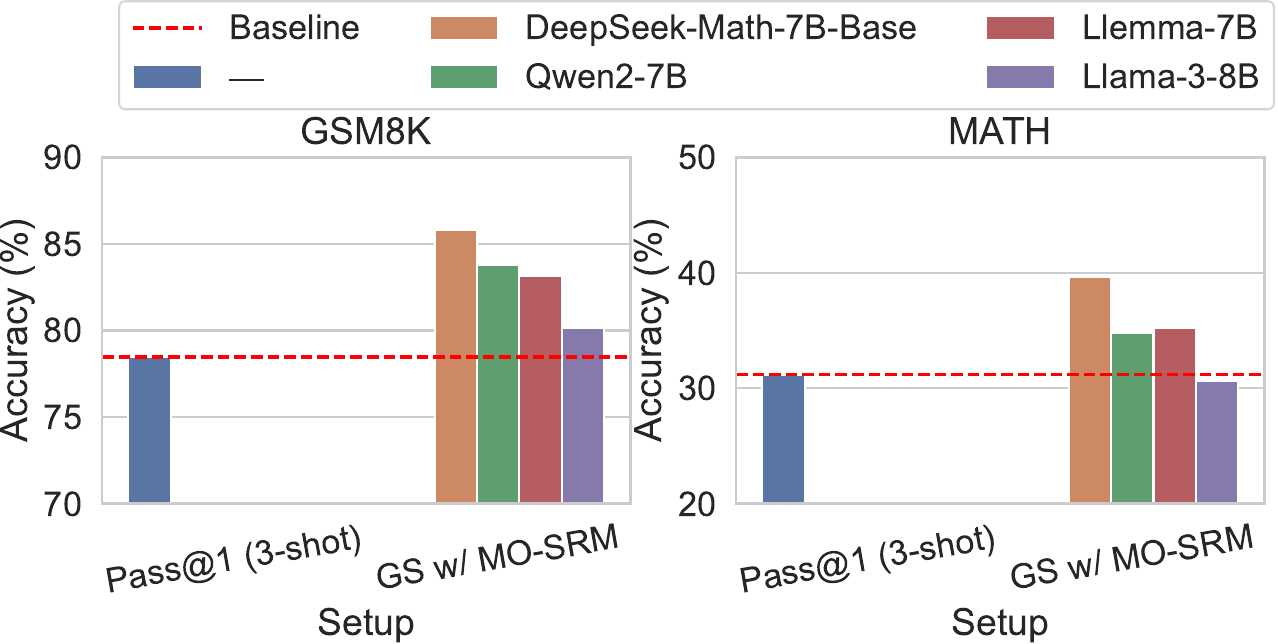}
  \caption{The performance of SRM is affected by the ability of the base model.}
  \label{fig:acx_vs_model}
\end{figure}

\subsubsection{Supervising a larger model} Despite being trained on preference data generated by a smaller model, the MO-SRM was able to effectively guide the reasoning process of a larger model and achieve substantial improvements (+2.58\% on GSM8K and +5.64\% on MATH) (\Cref{tab:llama3_70b}). This further illustrates the ability of the SRMs to focus exclusively on mathematical language.

\subsubsection{Effect of base models for MO-SRM}
The choice of SRM base models impacts performance (\Cref{fig:acx_vs_model}), while this effect doesn’t appear to be entirely related to the base model’s mathematical abilities. Despite its excellent mathematical capabilities, The surprising underperformance of Llama-3-8B compared to Llemma-7B \cite{azerbayev2023llemma}, Qwen2-7B, and DeekSeek-Math-7B-Base, suggests that factors beyond just original mathematical ability are at play. This might be due to the challenges in self-assessment or other reasons to be explored.

\begin{table}[H]
  \centering
  \begin{tabular}{llc}
    \toprule
    \textbf{Agent \& World Model}&  \multicolumn{2}{c}{\textbf{Accuracy}}\\
    \cmidrule{2-3}
    Llama-3-8B-Instruct&  \textbf{GSM8K}& \textbf{Accuracy}\\
    \midrule
    +BS w/ MO-SRM\footnotemark[1] & & \\
    \hspace{0.4cm}$B=1$, $c=5$& 85.82 & 39.64\\
    \hspace{0.4cm}$B=1$, $c=10$&  85.90& 40.06\\
    \hspace{0.4cm}$B=3$, $c=10$&  88.17& 40.24\\
    \bottomrule
  \end{tabular}
  \caption{Effect of $B$ and $c$ on beam search}
  \label{tab:effect_bc}
\end{table}

\subsubsection{Effect of $B$ and $c$ on beam search}
Increasing the beam size $B$ and the number of candidate count $c$ will slightly improve accuracy, but this improvement will eventually plateau, as shown in \Cref{tab:effect_bc}.

\section{Conclusion}
Our investigation into the role of natural language and mathematical expressions in step-level reward modeling reveals that natural language descriptions are not essential for the success of these models. Through extensive experiments, we demonstrated that reward models operating solely on mathematical expressions perform comparably to those that incorporate both natural language and math. Furthermore, the difficulty in training models to evaluate the coherence of natural language thought processes underscores the challenges LLMs face in capturing implicit logical structures through language alone. We also found that the coherence of logical structure inherent in mathematical expressions can be assessed by SRMs trained based on LLMs. Given the overhead of obtaining step-level rewards, these findings offer new insights for developing more efficient and targeted reward models by isolating the most impactful components of mathematical reasoning steps.

\section*{Acknowledgments}
This work was supported in part by National Key R\&D Program of China, under Grant No. 2022YFC3303600 and in part by Key Laboratory of Smart Education of Guangdong Higher Education Institutes, Jinan University (2022LSYS003).

\bibliography{aaai2025}

\onecolumn
\appendix
\setcounter{secnumdepth}{2}
\setcounter{figure}{0}
\setcounter{table}{0}
\renewcommand{\thefigure}{\thesection.\arabic{figure}}
\renewcommand{\thetable}{\thesection.\arabic{table}}
\section{Implementation Details}
\subsection{Prompts}\label{apdx:prompt}
\definecolor{HTMLCBE2B5}{HTML}{CBE2B5}
\begin{figure}[ht!]
  \begin{tcolorbox}[
      colframe=HTMLCBE2B5,  % 使用颜色代码指定边框颜色
      coltitle=black,           % 标题颜色
      title=System message (Agent),    % 标题文本
      % fonttitle=\itshape,       % 标题字体样式
      sharp corners,            % 直角边框
    ]
    You should act as a guide. You will break down the process into individual, understandable guidance step-by-step, each leading logically to the final result. I will follow your guidance by calculating the answer to each step with equations.\\\\
    \#\#\# Your response must meet the following requirements:\\
    1. Never say anything not related to the math problem.\\
    2. You should not include any calculations in your instruction as that is the student's work.\\
    3. If the current math problem is ready to be solved by following your next guidance, start it with ``Now you can answer the problem in this step.".\\
    4. If the final answer to the current math problem has been obtained, just say ``The math problem has been solved."
  \end{tcolorbox}
\end{figure}
\begin{figure}[ht!]
  \begin{tcolorbox}[
      colframe=HTMLCBE2B5,  % 使用颜色代码指定边框颜色
      coltitle=black,           % 标题颜色
      title=System message (World Model-GSM8K),    % 标题文本
      % fonttitle=\itshape,       % 标题字体样式
      sharp corners,            % 直角边框
    ]
    You are a student solving math problems under the instructions of the teacher. You should follow the step-by-step guidance posed by the teacher by calculating the answer of each step with equations until you deduce the final answer to the math problem.\\\\
    \#\#\# Your response must meet the following requirements:\\
    1. Never talk about anything not related to the math problem.\\
    2. Include the equation of this step.\\
    3. If the guidance starts with ``Now you can answer the problem in this step.", you must find the final answer to the problem in this step.\\
    4. End with ``The answer is" along with a single number to highlight the numerical (sub)answer (e.g. ``The answer is 42.").
  \end{tcolorbox}
\end{figure}
\begin{figure}[ht!]
  \begin{tcolorbox}[
      colframe=HTMLCBE2B5,  % 使用颜色代码指定边框颜色
      coltitle=black,           % 标题颜色
      title=System message (World Model-MATH),    % 标题文本
      % fonttitle=\itshape,       % 标题字体样式
      sharp corners,            % 直角边框
    ]
    You are a student solving math problems under the instructions of the teacher. You should follow the step-by-step guidance posed by the teacher by calculating the answer of each step with equations until you deduce the final answer to the math problem.\\\\
    \#\#\# Your response must meet the following requirements:\\
    1. Include the equation of this step.\\
    2. If the subquestion is started with start it with ``Now you can answer the problem in this step.\", you must find the final answer to the problem in this step.\\
    3. You must use the LaTeX code ``\\boxed{}" to highlight the final answer to the problem. (e.g. ``$(9+1)^3 = 10^3 = \boxed{1000}$").
  \end{tcolorbox}
\end{figure}
\subsection{Hyperparameters}\label{apdx:hyperparam}
\subsubsection{MCTS} The hyperparameters of MCTS are shown in \Cref{tab:mcts_param}.
\begin{table}[H]
  \centering
  \begin{tabular}{cc}
    \toprule
    Hyperparameter      & Value \\
    \midrule
    $n$ (n\_candidates) & 6     \\
    depth\_limit        & 8     \\
    $w_{exp}$           & 1.0   \\
    temperature (agent) & 1.3   \\
    temperature (world) & 0.7   \\
    n\_iteration        & 500   \\
    \bottomrule
  \end{tabular}
  \caption{Hyperparameters of MCTS}
  \label{tab:mcts_param}
\end{table}
\subsubsection{Step-Level Reward Modeling} The hyperparameters for step-level reward modeling are shown in \Cref{tab:srm_param}.
\begin{table}[H]
  \centering
  \begin{tabular}{cc}
    \toprule
    Hyperparameter                  & Value   \\
    \midrule
    n\_instances                    & 2       \\
    gpus\_per\_instance             & 8       \\
    per\_device\_train\_batch\_size & 16      \\
    gradient\_accumulation\_steps   & 2       \\
    num\_train\_epochs              & 2       \\
    warmup\_ratio                   & 0.03    \\
    learning\_rate                  & 1.41e-5 \\
    weight\_decay                   & 0.1     \\
    \bottomrule
  \end{tabular}
  \caption{Hyperparameters of MCTS}
  \label{tab:srm_param}
\end{table}

\subsubsection{BS w/ SRM} The hyperparameters for BS w/ SRM are shown in \Cref{tab:bssrm_param}.
\begin{table}[H]
  \centering
  \begin{tabular}{cc}
    \toprule
    Hyperparameter      & Value \\
    \midrule
    $n$ (n\_candidates) & 5 or 10     \\
    beam\_size          & 1 or 3     \\
    temperature (agent) & 0.7   \\
    temperature (world) & 0.0   \\
    \bottomrule
  \end{tabular}
  \caption{Hyperparameters of MCTS}
  \label{tab:bssrm_param}
\end{table}
\subsection{Example}
\subsubsection{Greedy Search Supervised by an SRM}\label{apdx:exmple_case}
\begin{figure}[H]
  \centering
  \includegraphics[width=\linewidth]{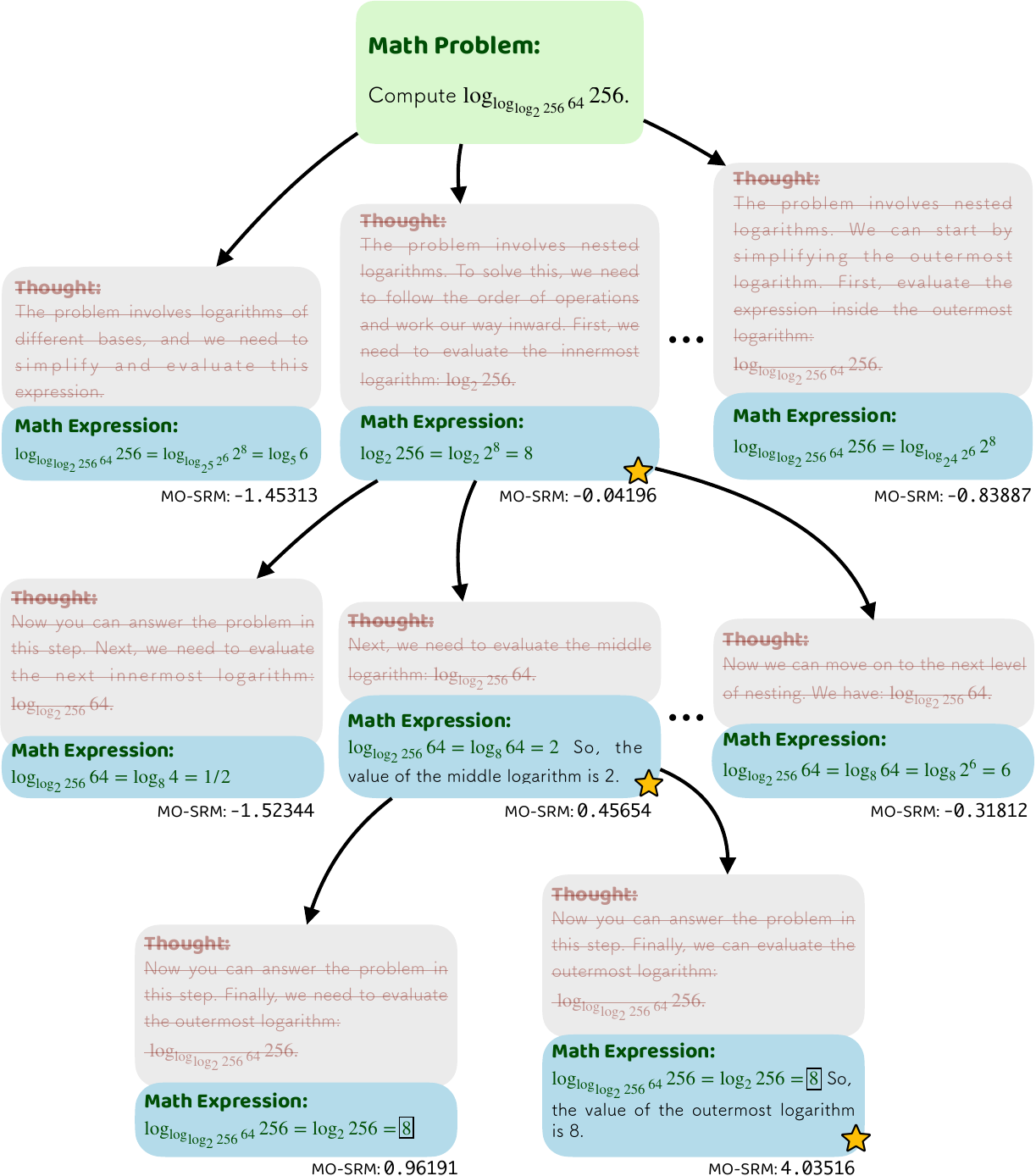}
  \caption{An example of a case where the MO-SRM is used to supervise the GS.}
  \label{fig:example_case}
\end{figure}

For a better understanding of our definition of the mathematical reasoning process supervised by the SRMs, we provide an example (\Cref{fig:example_case}) of a greedy search, where the rewards are from the MO-SRM.

\subsection{Addtional Results}
\subsection{Tendency of encouraging shorter paths}
\begin{figure}[H]
  \centering
  \includegraphics[width=0.6\linewidth]{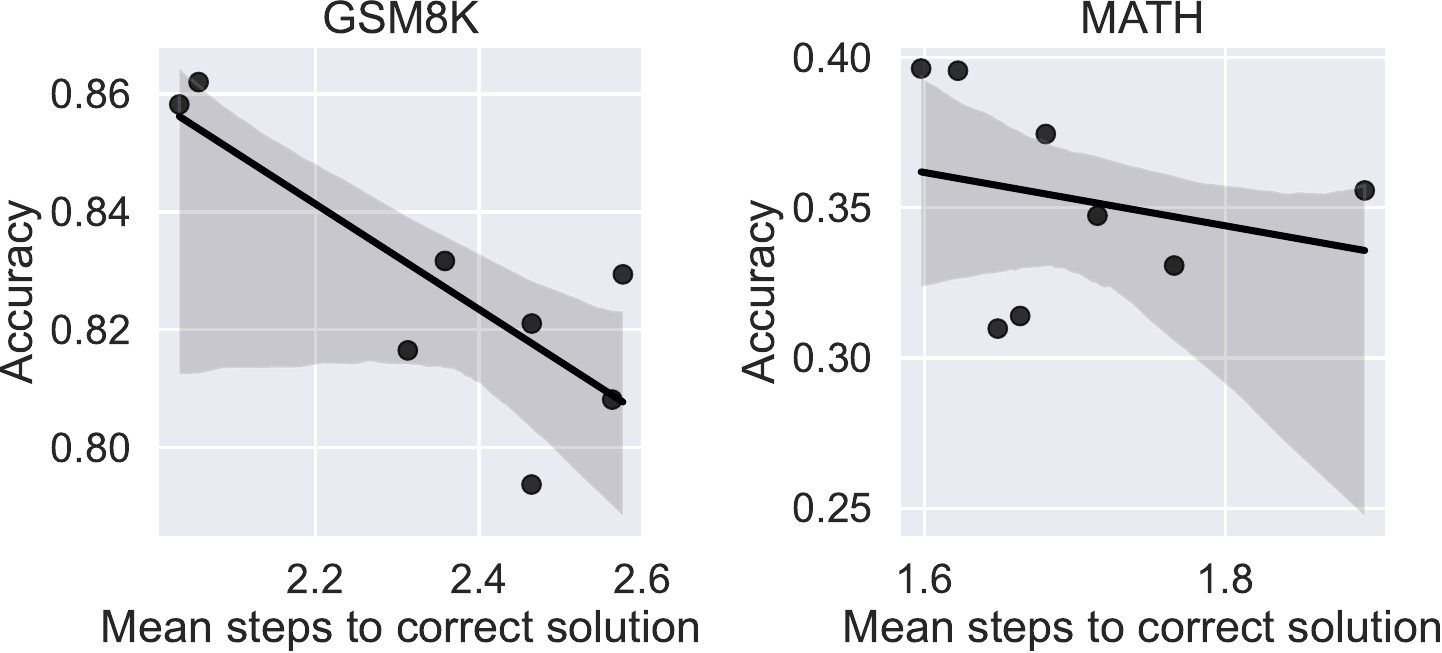}
  \caption{Accuracy v.s. mean steps to correct solutions. Fewer steps to correct solutions tend to have higher accuracy.}
  \label{fig:step_reward}
\end{figure}

We observed that the greedy search with the SRMs tends to encourage shorter reasoning paths, although the MCTS itself does not explicitly include the path length as a preference. (\Cref{fig:step_reward}) This observation is due to the insufficient exploitation of the MCTS process, but we need further investigation to confirm this proposition in future studies.

\end{document}